%% file: iclr2021_conference.tex
\documentclass{article} 
\usepackage{iclr2021_conference,times}

\usepackage{xspace}
\usepackage{hyperref}
\usepackage{url}
\usepackage{subfigure}
\usepackage{times}
\usepackage{adjustbox}      
\usepackage{epsfig}
\usepackage{graphicx}
\usepackage{amsmath}
\usepackage{amssymb}
\usepackage{booktabs}

\input{tex/abbrev}
\input{tex/defn}

\title{Few-shot learning via tensor hallucination}

\author{
Michalis Lazarou$^1$ \ \ \ \ Yannis Avrithis$^2$\ \ \ \ Tania Stathaki$^1$\\
$^1$Imperial College London\\
$^2$Inria, Univ Rennes, CNRS, IRISA\\
}

\iclrfinalcopy 

\begin{document}

\maketitle

\begin{abstract}
Few-shot classification addresses the challenge of classifying examples given only limited labeled data. A powerful approach is to go beyond data augmentation, towards data synthesis. However, most of data augmentation/synthesis methods for few-shot classification are overly complex and sophisticated, \eg training a wGAN with multiple regularizers or training a network to transfer latent diversities from known to novel classes. We make two contributions, namely we show that: (1) using a simple loss function is more than enough for training a feature generator in the few-shot setting; and (2) learning to generate tensor features instead of vector features is superior. Extensive experiments on \emph{mini}Imagenet, CUB and CIFAR-FS datasets show that our method sets a new state of the art, outperforming more sophisticated few-shot data augmentation methods.
\end{abstract}

\input{tex/intro}
\input{tex/method}
\input{tex/exp}

\input{tex/conclusion}

 


 
\newpage
\bibliography{iclr2021_conference}
\bibliographystyle{iclr2021_conference}

 
\newpage
\input{tex/app}
\end{document}

%% file: tex/abbrev.tex
\newcommand{\alert}[1]{{\color{red}{#1}}}

\newcommand{\eq}[1]{(\ref{eq:#1})}

\newcommand{\Th}[1]{\textsc{#1}}

\newcommand{\mc}[2]{\multicolumn{#1}{c}{#2}}
\newcommand{\tb}[1]{\textbf{#1}}

\newcommand{\red}[1]{{\color{red}{#1}}}

\newcommand{\citeme}[1]{\red{[XX]}}
\newcommand{\refme}[1]{\red{(XX)}}

\newcommand{\real}{\mathbb{R}}

\newcommand{\normal}{\mathcal{N}}

\newcommand{\defn}{\mathrel{:=}}

\newcommand{\norm}[1]{\left\|{#1}\right\|}

\newcommand{\cX}{\mathcal{X}}

\newcommand{\vzero}{\mathbf{0}}

\makeatletter
\newcommand*\bdot{\mathpalette\bdot@{.7}}
\newcommand*\bdot@[2]{\mathbin{\vcenter{\hbox{\scalebox{#2}{$\m@th#1\bullet$}}}}}
\makeatother

\makeatletter
\DeclareRobustCommand\onedot{\futurelet\@let@token\@onedot}
\def\@onedot{\ifx\@let@token.\else.\null\fi\xspace}

\def\eg{\emph{e.g}\onedot}

\makeatother

%% file: tex/defn.tex
\newcommand{\base}{\mathrm{base}}
\newcommand{\novel}{\mathrm{novel}}
\newcommand{\CE}{\mathrm{CE}}
\newcommand{\KL}{\mathrm{KL}}
\newcommand{\KD}{\mathrm{KD}}
\newcommand{\hal}{\mathrm{hal}}

%% file: tex/intro.tex
\section{Introduction}
\label{intro}
Deep learning continuously keeps improving the state of the art in multiple different fields, such as natural language understanding~\cite{language} and computer vision~\cite{imagenet}. However, even though the success of deep learning models is undeniable, a fundamental limitation is their dependence on large amounts of labeled data. This limitation inhibits the application of state of the art deep learning methods to real-world problems, where the cost of annotating data is high and data can be scarce, \eg rare species classification.

To address this limitation, \emph{few-shot learning} has attracted significant interest in recent years. 
One of the most common lines of research is \emph{meta-learning}, 
where training episodes
mimic a few-shot task by having a small number of classes and a limited number of examples per class. Meta-learning approaches can further be partitioned in \emph{optimization-based}, learning to update the learner's meta-parameters~\cite{MAML, CAVIA, reptile, ravilstm}, \emph{metric-based}, learning a discriminative embedding space where novel examples are easy to classify~\cite{siamese, prototypical, relationnet, matchingNets} and \emph{model-based}, depending on specific model architectures to learn how to update the learner's parameters effectively~\cite{santoro, metaNet}.

Beyond meta-learning, other approaches include leveraging the \emph{manifold structure} of the data, by label propagation, embedding propagation or graph neural networks~\cite{fewshotGNN1, fewshotGNN2, tpn, ilpc}; and \emph{domain adaptation}, reducing the domain shift between source and target domains~\cite{domain_one, l2cluster}. Another line of research is \emph{data augmentation}, addressing data deficiency by augmenting the few-shot training dataset with extra examples in the image space~\cite{Ideme, metaGAN} and in the feature space~\cite{semanticAugmentation, AFHN, VIFSL}. Such methods go beyond standard augmentation~\cite{imagenet} towards \emph{synthetic data generation} and \emph{hallucination}, achieving a greater extent of diversity. 

Our work falls into the category of data augmentation in the feature space. We show that using a simple loss function to train a feature hallucinator can outperform other state of the art few-shot data augmentation methods that use more complex and sophisticated generation methods, such as wGAN~\cite{AFHN}, VAE~\cite{VIFSL} and networks trained to transfer example diversity~\cite{DTN}. Also, to the best of our knowledge, we are the first to propose generating \emph{tensor features} instead of vector features in the few-shot setting.

%% file: tex/method.tex
\section{Method}
\label{sec:method}

\subsection{Problem formulation}
\label{sec:problem}

We are given a labeled dataset $D_{\base} \defn \{(x_i, y_i)\}_{i=1}^I$, with each example $x_i$ having a label $y_i$ in one of the classes in $C_{\base}$. This dataset is used to learn a parametrized mapping $f_\theta: \cX \to \real^{d \times h \times w}$ from an input image space $\cX$ to a \emph{feature} or \emph{embedding} space, where \emph{feature tensors} have $d$ dimensions (channels) and spatial resolution $h \times w$ (height $\times$ width).

The knowledge acquired at representation learning is used to solve \emph{novel tasks}, assuming access to a dataset $D_{\novel}$, with each example being associated with one of the classes in $C_{\novel}$, where $C_{\novel}$ is disjoint from $C_{\base}$. In \emph{few-shot classification}~\cite{matchingNets}, a novel task is defined by sampling a \emph{support set} $S$ from $D_{\novel}$, consisting of $N$ classes with $K$ labeled examples per class, for a total of $L \defn NK$ examples. Given the mapping $f_\theta$ and the support set $S$, the problem is to learn an $N$-way classifier that makes predictions on unlabeled \emph{queries}, also sampled from $D_{\novel}$. Queries are treated independently of each other. This is referred to as \emph{inductive inference}.


\subsection{Representation learning}
\label{sec:representation}

The goal of \emph{representation learning} is to learn the embedding function $f_\theta$ that can be applied to $D_{\novel}$ to extract embeddings and solve novel tasks. We use $f_\theta$ followed by \emph{global average pooling} (GAP) and a parametric \emph{base classifier} $c_\phi$ to learn the representation. We denote by $\bar{f}_\theta: \cX \to \real^d$ the composition of $f_\theta$ and GAP. We follow the two-stage regime by~\cite{rfs} to train our embedding model. In the first stage, we train $f_\theta$ on $D_{\base}$ using standard cross-entropy loss $L_{\CE}$:
\begin{equation}
    L(D_{\base}; \theta, \phi) \defn
		\sum_{i=1}^I L_{\CE}(c_{\phi}(\bar{f}_\theta(x_i)), y_i) + R(\phi),
\label{eq:cls}
\end{equation}
where $R$ is a regularization term. In the second stage, we adopt a \emph{self-distillation} process: The embedding model $f_\theta$ and classifier $c_\phi$ from the first stage serve as the teacher and we distill their knowledge to a new student model $f_{\theta'}$ and classifier $c_{\phi'}$, with identical architecture. The student is trained using a linear combination of the standard cross-entropy loss, as in stage one, and the Kullback-Leibler (KL) divergence between the student and teacher predictions:
\begin{equation}
    L_{\KD}(D_{\base}; \theta', \phi') \defn
		\alpha L(D_{\base}; \theta', \phi') +
		\beta \KL(c_{\phi'}(\bar{f}_{\theta'}(x_i)), c_\phi(\bar{f}_{\theta}(x_i))),
\label{eq:kl}
\end{equation}
where $\alpha$ and $\beta$ are scalar weights and $\theta, \phi$ are fixed.


\subsection{Feature tensor hallucinator}

All existing feature hallucination methods are trained using vector features, losing significant spatial and structural information. By contrast, our hallucinator is trained on the tensor features before global average pooling and generates tensor features as well. In particular, we use the student model $f_{\theta'}: \cX \to \real^{d \times h \times w}$, pre-trained using \eq{kl}, as our embedding network to train our tensor feature hallucinator. The hallucinator consists of two networks: a \emph{conditioner} network $h$ and a \emph{generator} network $g$. The conditioner aids the generator in generating class-conditional examples. Given a set $\{x_i^j\}_{i=1}^K$ of examples associated with class $j$ for $j=1,\dots,N$, conditioning is based on the \emph{prototype tensor} $p_j \in \real^{d \times h \times w}$ of each class $j$,
\begin{equation}
    p_j \defn \frac{1}{K} \sum_{i=1}^{K} f_{\theta'}(x_i^j).
\label{eq:proto}
\end{equation}
The conditioner $h: \real^{d \times h \times w} \to \real^{d'}$ maps the prototype tensor to the \emph{class-conditional vector} $s_j \defn h(p_j) \in \real^{d'}$. The generator $g: \real^{k+d'} \to \real^{d \times h \times w}$ takes as input this vector as well as a $k$-dimensional sample $z \sim \normal(\vzero,I_k)$ from a standard normal distribution and generates a \emph{class-conditional tensor feature} $g(z; s_j) \in \real^{d \times h \times w}$ for class $j$.


\subsection{Training the hallucinator}

We train our hallucinator using a meta-training regime, similar to \cite{AFHN, semanticAugmentation, deltaencoder}. At every iteration, we sample a new episode by randomly sampling $N$ classes and $K$ examples $X_j \defn \{x_i^j\}_{i=1}^K$ for each class $j$ from $D_{\base}$. We obtain the prototype tensor $p_j$ for each class $j$ by~\eq{proto} and the class-conditional vector $s_j \defn h(p_j)$ by the conditioner $h$. For each class $j$, we draw $M$ samples $\{z_m\}_{m=1}^M$ from a normal distribution $\normal(\vzero,I_k)$ and we generate $M$ class-conditional tensor features $\{g(z_m; s_j)\}_{m=1}^M$ using the generator $g$. We train our hallucinator $\{h,g\}$ on the episode data $X \defn \{X_j\}_{j=1}^N$ by minimizing the \emph{mean squared error} (MSE) of generated class-conditional tensor features of class $j$ to the corresponding class prototype $p_j$:
\begin{equation}
	L_{\hal}(X; h, g) = \frac{1}{MN} \sum_{j=1}^N \sum_{m=1}^M \norm{g(z_m; h(p_j)) - p_j}^2.
\label{eq:mse}
\end{equation}


\subsection{Inference}

At inference, we are given a few-shot task with a support set $S$ of $N$ classes with $K$ examples $S_j \defn \{x_i^j\}_{i=1}^K$ for each class $j$. We compute the tensor feature $f_{\theta'}(x_i^j) \in \real^{d \times h \times w}$ of each example using our trained backbone network $f_{\theta'}$ and obtain the prototype $p_j$ of each class $j$ by~\eq{proto}. Then, using our trained tensor feature hallucinator $\{h,g\}$, we generate $M$ class-conditional tensor features $G_j \defn \{g(z_m; h(p_j))\}_{m=1}^M$ for each class $j$, also in $\real^{d \times h \times w}$, where $z_m$ are drawn from $\normal(\vzero,I_k)$. We augment the support features $f_{\theta'}(S_j)$ with the generated features $G_j$, resulting in $K+M$ labeled tensor features per class in total. We now apply GAP to those tensor features and obtain new, \emph{vector class prototypes} in $\real^d$. Finally, we also apply GAP to the query tensor features and classify each query to the class of the nearest prototype.

%% file: tex/exp.tex
\section{Experiments}
\label{experiments}

\newcommand{\ci}[1]{{\tiny $\pm$#1}}
\newcommand{\cip}{\phantom{\ci{0.00}}}
\newcommand{\cim}{\ci{\alert{0.00}}}

 
\subsection{Setup}

\paragraph{Datasets}

We carry out experiments on three commonly used few-shot classication benchmark datasets: \emph{mini}Imagenet, CUB and CIFAR-FS. Further details are provided in \autoref{sec:datasets}.

\paragraph{Tasks} 

We consider $N$-way, $K$-shot classification tasks with $N = 5$ randomly sampled novel classes and $K \in \{1, 5\}$ examples drawn at random per class as support set $S$, that is, $L = 5K$ examples in total. For the query set $Q$, we draw $15$ additional examples per class, that is, $75$ examples in total, which is the most common choice~\cite{tpn, learning2selftrain, transmatch}.

\paragraph{Competitors}

We compare our method with state-of-the-art data augmentation methods for few-shot learning, including MetaGAN~\cite{metaGAN}, $\Delta$-encoder~\cite{deltaencoder}, salient network (SalNet)~\cite{salnet}, diversity transfer network~(DTN) \cite{DTN}, dual TriNet~\cite{semanticAugmentation}, image deformation meta-network (IDeMe-Net)~\cite{Ideme}, adversarial feature hallucination network (AFHN)~\cite{AFHN} and variational inference network (VI-Net)~\cite{VIFSL}.


\paragraph{Networks}

Many recent competitors \cite{semanticAugmentation, Ideme, AFHN, VIFSL} use ResNet-18 as backbone embedding model. To perform as fair comparison as possible, we use the same backbone.

Our \emph{tensor feature hallucinator} (TFH) consists of a conditioner network and a generator network. The \emph{conditioner} $h: \real^{d \times h \times w} \to \real^{d'}$ consists of two convolutional layers with a ReLU activation in-between, followed by flattening and a fully-connected layer. The \emph{generator} $g: \real^{k+d'} \to \real^{d \times h \times w}$ consists of concatenation of $z$ and $s_j$ into $(z; s_j) \in \real^{k+d'}$, followed by reshaping to $(k+d') \times 1 \times 1$ and three transpose-convolutional layers with ReLU activation functions in-between and a sigmoid function in the end. More details are provided in \autoref{sec:impl}.


We also provide an improved solution, called TFH-ft, where our tensor feature hallucinator is \emph{fine-tuned} on novel-class support examples at inference.

\paragraph{Baselines}

To validate the benefit of generating tensor features, we also implement a \emph{vector feature hallucinator} (VFH), where we use $\bar{f}_{\theta'}: \cX \to \real^d$ including GAP~\eq{kl} as embedding model. In this case, the \emph{conditioner} $h: \real^d \to \real^{d'}$ consists of two fully-connected layers with a ReLU activation in-between. The \emph{generator} $g: \real^{k+d'} \to \real^d$ also consists of two fully-connected layers with a ReLU activation in-between and a sigmoid function in the end.

Finally, we experiment with baselines consisting of the embedding network $f_\theta$~\eq{cls} or $f_{\theta'}$~\eq{kl} at representation learning and a prototypical classifier at inference, without feature hallucination. We refer to them as Baseline~\eq{cls} and Baseline-KD~\eq{kl} respectively.

\begin{table*}
\small
\centering
\begin{adjustbox}{width=1\textwidth}
\setlength\tabcolsep{4pt}
\begin{tabular}{lccccccc} \toprule
\Th{Method}                                      & \Th{Backbone} & \mc{2}{\Th{\emph{mini}ImageNet}}          & \mc{2}{\Th{CUB}}                          & \mc{2}{\Th{CIFAR-FS}}                     \\  
                                                 &               & 1-shot              & 5-shot              & 1-shot              & 5-shot              & 1-shot              & 5-shot              \\ \midrule
MetaGAN \cite{metaGAN}                           & ConvNet-4     & 52.71\ci{0.64}      & 68.63\ci{0.67}      & --                  & --                  & --                  & --                  \\
$\Delta$-Encoder$\dagger$ \cite{deltaencoder}    & VGG-16        & 59.90\cip           & 69.70\cip           & 69.80\ci{0.46}      & 82.60\ci{0.35}      & 66.70\cip           & 79.80\cip           \\
SalNet \cite{salnet}                             & ResNet-101    & 62.22\ci{0.87}      & 77.95\ci{0.65}      & --                  & --                  & --                  & --                  \\
DTN \cite{DTN}                                   & Resnet-12     & 63.45\ci{0.86}      & 77.91\ci{0.62}      & 72.00\cip           & 85.10\cip           & 71.50\cip           & 82.80\cip           \\ \midrule
Dual TriNet \cite{semanticAugmentation}          & ResNet-18     & 58.80\ci{1.37}      & 76.71\ci{0.69}      & 69.61\cip           & 84.10\cip           & 63.41\ci{0.64}      & 78.43\ci{0.64}      \\ 
IDeMe-Net \cite{Ideme}                           & ResNet-18     & 59.14\ci{0.86}      & 74.63\ci{0.74}      & --                  & --                  & --                  & --                  \\ 
AFHN \cite{AFHN}                                 & ResNet-18     & 62.38\ci{0.72}      & 78.16\ci{0.56}      & 70.53\ci{1.01}      & 83.95\ci{0.63}      & 68.32\ci{0.93}      & 81.45\ci{0.87}      \\
VI-Net \cite{VIFSL}                              & ResNet-18     & 61.05\cip           & 78.60\cip           & 74.76\cip           & 86.84\cip           & --                  & --                  \\ \midrule
Baseline~\eq{cls}                                & ResNet-18     & 56.81\ci{0.81}      & 78.31\ci{0.59}      & 67.14\ci{0.89}      & 86.22\ci{0.50}      & 65.71\ci{0.95}      & 84.68\ci{0.61}      \\
Baseline-KD~\eq{kl}                              & ResNet-18     & 59.62\ci{0.85}      & 79.64\ci{0.62}      & 70.85\ci{0.90}      & 87.64\ci{0.48}      & 69.15\ci{0.94}      & 85.89\ci{0.59}      \\
VFH (ours)                                       & ResNet-18     & 61.92\ci{0.85}      & 77.02\ci{0.64}      & 75.25\ci{0.86}      & 87.96\ci{0.48}      & 72.60\ci{0.93}      & 84.26\ci{0.67}      \\
\tb{TFH (ours)}                                  & ResNet-18     & \tb{64.25}\ci{0.85} & 80.10\ci{0.61}      & \tb{75.83}\ci{0.91} & 88.17\ci{0.48}      & 73.88\ci{0.87}      & 85.92\ci{0.61}      \\
\tb{TFH-ft (ours)}                               & ResNet-18     & 63.92\ci{0.86}      & \tb{80.41}\ci{0.60} & 75.39\ci{0.86}      & \tb{88.72}\ci{0.47} & \tb{73.89}\ci{0.88} & \tb{87.15}\ci{0.58} \\
\bottomrule
\end{tabular}
\end{adjustbox}
\vspace{6pt}
\caption{Comparison of our proposed method variants and baselines to state of the art on few-shot classification datasets. $\dagger$: Delta-encoder uses VGG-16 backbone for \emph{mini}ImageNet and CIFAR-FS and ResNet-18 for CUB. Baseline~\eq{cls}, Baseline-KD~\eq{kl}: prototypical classifier at inference, no feature generation. VFH: our vector feature hallucinator; TFH: our tensor feature hallucinator; TFH-ft: our tensor feature hallucinator followed by fine-tuning at inference.}
\label{tab:sota}
\end{table*}

\subsection{Results}

\autoref{tab:sota} compares our method with the state of the art. Most important are comparisons with \cite{semanticAugmentation, Ideme, AFHN, VIFSL}, which use the same backbone, ResNet-18. Our tensor feature hallucinator provides new state of the art performance in all datasets and all settings, outperforming all competing few-shot data augmentation methods. \emph{Fine-tuning} at inference is mostly beneficial, especially at 5-shot tasks. This is expected, as more data means less risk of overfitting. It is clear that the \emph{tensor feature hallucinator} is superior to the vector feature hallucinator, while the latter is still very competitive. \emph{Self-distillation} also provides a significant boost of performance in all experiments.

%% file: tex/conclusion.tex
\section{Conclusion}
\label{conclusion}
Our solution is conceptually simple and improves the state of the art of data augmentation methods in the few-shot learning setting. We provided experimental evidence showing that using a simple loss function and exploiting the structural properties of tensors can provide significant improvement in performance. Notably, the importance of using tensor features is evident through comparison with vector features, which are unable to achieve similar performance. Potential future directions include investigating the performance of our method with different backbone architectures and other experimental settings beyond few-shot learning.

%% file: tex/app.tex
\appendix

\section{Appendix}

\subsection{Dataset details}
\label{sec:datasets}

\paragraph{\emph{mini}ImageNet}

This is a widely used few-shot image classification dataset~\cite{matchingNets, ravilstm}. It contains 100 randomly sampled classes from ImageNet~\cite{imagenet}. These 100 classes are split into 64 training (base) classes, 16 validation (novel) classes and 20 test (novel) classes. Each class contains 600 examples (images). We follow the commonly used split provided by~\cite{ravilstm}.

\paragraph{CUB}

This is a fine-grained classification dataset consisting of 200 classes, each corresponding to a bird species. We follow the split defined by~\cite{closerlook, fewshotCUB}, with 100 training, 50 validation and 50 test classes. 

\paragraph{CIFAR-FS}

This dataset is derived from CIFAR-100~\cite{cifar100db}, consisting of 100 classes with 600 examples per class. We follow the split provided by~\cite{closerlook}, with 64 training, 16 validation and 20 test classes.

All images from all datasets are resized to $224 \times 224$ in a similar way to other data augmentation methods \cite{AFHN, semanticAugmentation, Ideme, VIFSL}

 
\subsection{Implementation details}
\label{sec:impl}

Our implementation is based on PyTorch \cite{pytorch}.

\paragraph{Networks}

In our \emph{tensor feature hallucinator} (TFH), the embedding dimension is $d = 512$ and the resolution $h \times w$ is $7 \times 7$.

The convolutional layers of the \emph{conditioner} use kernels of size $3 \times 3$ and stride 1 and in the input layer we also use padding 1. The channel dimensions are 512 and 256 for the first and second convolutional layers respectively. The dimension of the \emph{class-conditional vector} is set to $d' = 1024$. The tensor dimensions of all conditioner layers are $[512 \times 7 \times 7]$, $[256 \times 5 \times 5]$, $[6400]$ (flattening) and $[1024]$.

All three transpose-convolutional layers of the \emph{generator} use kernels of size $3 \times 3$, stride 1 and 512 channels. The dimension of $z \sim \normal(\vzero,I_{k})$ is $k = 1024$. The tensor dimensions of all generator layers are $[2048 \times 1 \times 1]$, $[512 \times 3 \times 3]$, $[512 \times 5 \times 5]$, and $[512 \times 7 \times 7]$.

In our \emph{vector feature hallucinator} (VFH), the dimensions of the \emph{class-conditional vector} as well as the hidden layers of both the \emph{conditioner} and the \emph{generator} are all set to 512.

 
\paragraph{Training}

For the \emph{embedding model}, similarly to \cite{rfs}, we use SGD optimizer with learning rate 0.05, momentum 0.9 and weight decay 0.0005. For data augmentation, as in \cite{metaridge}, we adopt random crop, color jittering, and horizontal flip.

The \emph{tensor feature hallucinator} is trained in a meta-training regime with $N = 5$ classes, $K = 20$ examples per class and  generation of $M = 50$ class-conditioned examples in every task. We use Adam optimizer with initial learning rate $10^{-5}$, decaying by half at every 10 epochs. We train for 50 epochs, where each epoch consists of 600 randomly sampled few-shot learning tasks. At test time, we find that generating more class-conditioned examples improves the accuracy, therefore we generate $M = 500$ class-conditioned examples.

Our TFH-ft version uses the novel-class support examples to fine-tune all of its parameters. In the fine-tuning stage, we use exactly the same loss function as in the hallucinator training phase~\eq{mse} and we fine-tune for 10 steps using Adam optimizer and learning rate of $10^{-5}$.